%% file: main.tex
\documentclass[letterpaper,10pt,conference]{ieeeconf}

\IEEEoverridecommandlockouts

\makeatletter
\let\NAT@parse\undefined
\makeatother
\usepackage[numbers,sort&compress]{natbib}
\usepackage{graphicx}
\usepackage{booktabs}
\usepackage{amsmath}
\usepackage{amsfonts}
\usepackage{amssymb}
\usepackage{bm}
\usepackage{xspace}
\usepackage[table]{xcolor}
\usepackage{multirow}
\usepackage{tabularx}
\usepackage{array}
\usepackage{subcaption}
\usepackage{placeins}
\usepackage{pifont}
\usepackage{makecell}
\usepackage{tikz}
\usepackage{pgfplots}
\usepackage[normalem]{ulem}
\usepackage{microtype}
\usepackage{url}
\usepgfplotslibrary{groupplots}

\pgfplotsset{compat=1.18}

\newcommand{\modelname}{\textbf{\texttt{SPELL}}\xspace}

\definecolor{deletedgray}{gray}{0.25}
\definecolor{picolor}{RGB}{72,120,166}
\definecolor{nncolor}{RGB}{85,168,104}
\definecolor{oftcolor}{RGB}{221,132,82}
\definecolor{ourscolor}{RGB}{196,78,82}
\definecolor{oraclecolor}{RGB}{90,90,90}

\title{\LARGE \bf
Long-horizon Compositional Manipulation\\via Meta-Network Dynamic Adaptation
}

\author{
Christian Bianchi$^{1,2,*,\dagger}$,
Mehdi Amlal$^{2,*}$,
Alessio Palma$^{2}$,
Matteo Gioia$^{2}$,
Edoardo Ensoli$^{2}$,\\
Alessio Sampieri$^{1}$,
Andrea Roberti$^{3}$,
Fabio Galasso$^{1,2}$,
Luca Franco$^{1}$\\[1ex]
\small
$^{1}$ItalAI
\quad
$^{2}$Sapienza University of Rome, Italy
\quad
$^{3}$University of Verona, Italy\\
$^{*}$Equal contribution
\quad
$^{\dagger}$Corresponding author}

\begin{document}
\bstctlcite{BSTcontrol}
\maketitle
\thispagestyle{empty}
\pagestyle{empty}

\begin{abstract}
Vision-language-action (VLA) policies have greatly expanded the range of robotic manipulation behaviors learned from large-scale imitation data. However, adapting them to new long-horizon tasks remains challenging, as a single policy must identify, sequence, and execute multiple skills within a novel task composition. We introduce SPELL, a Skill-level Parameter Encoding framework for Long-horizon Learning that addresses this problem by composing task-specific adaptations directly in weight space. SPELL starts from LoRA experts trained on individual skills and learns a meta-network that, given a language instruction and visual demonstration, generates the adapters required for the skills composing a new task. During execution, SPELL dynamically activates the appropriate skill adapter as the task progresses, enabling adaptation to unseen compositions without target-task action labels or test-time gradient updates. We evaluate SPELL on held-out task compositions in CALVIN and LIBERO-10, on state-dependent compositions in Isaac Sim, and on a real-world Franka Emika Panda setup. Across these settings, SPELL improves long-horizon execution over monolithic policies and non-compositional adapter generation, showing that skill-level weight-space composition provides an effective mechanism for reusing learned expertise in novel task sequences.
\end{abstract}

\input{sections/1_introduction}
\input{sections/2_literature}
\input{sections/3_method}

\input{sections/4_experiments}
\input{sections/6_lim_con}
\FloatBarrier

\bibliographystyle{IEEEtran}
\bibliography{references}

\end{document}

%% file: sections/1_introduction.tex
\section{Introduction}

Long-horizon robotic manipulation is inherently \textit{compositional}: a robot must be able to reuse known skills, adapt them to new contexts or objects, and arrange them in the correct order. Models such as the VLA-based policies in
\citep{black2026pi0visionlanguageactionflowmodel}, \citep{kim2025openvla} or \citep{reuss2025flowerdemocratizinggeneralistrobot} have demonstrated that they can reach impressive performance on low-level manipulation tasks and excel at acquiring large amounts of basic abilities.
Despite this progress, existing VLA policies still struggle to solve new long-horizon tasks. By design, VLAs acquire skills as entangled behaviors inside a single parameter set, have to implicitly recover which skills are needed and understand when to invoke them, all while preserving execution fidelity across skill boundaries. This limitation becomes especially visible in long-horizon tasks \cite{zhang2024vlabenchlargescalebenchmarklanguageconditioned, han2025robocerebralargescalebenchmarklonghorizon}, out-of-domain settings \cite{pumacay2024colosseumbenchmarkevaluatinggeneralization, zhang2025vlaarenaopensourceframeworkbenchmarking}, and novel combinations of existing skills \cite{zhou2025liberoprorobustfairevaluation}.

\begin{figure*}[ht] 
\centering
\includegraphics[width=0.9\linewidth]{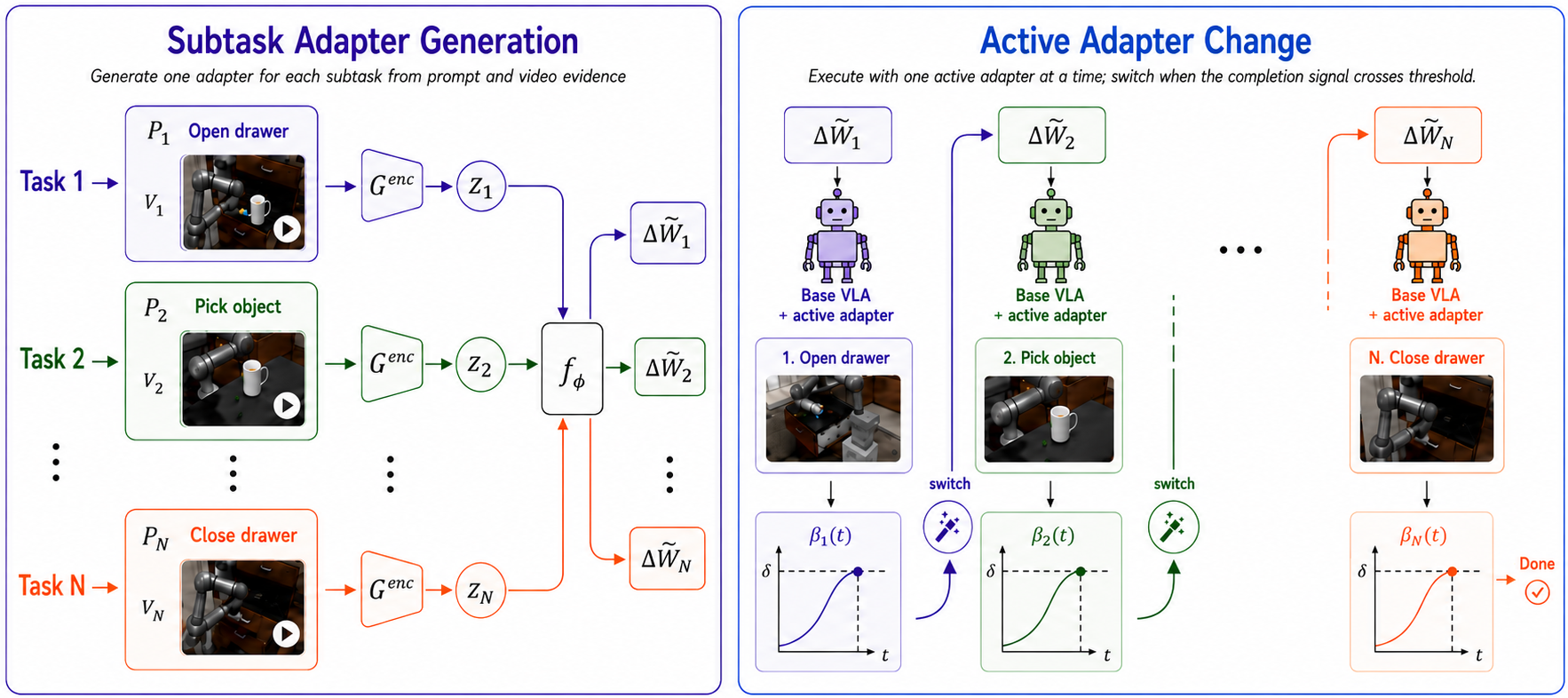}
\caption{
\textbf{\modelname overview.}
For subtask $n$, task evidence $(p_n,v_n)$ is encoded into $z_n$, from which the meta-network $f_\phi$ generates the skill-level adapter $\Delta\widetilde{W}_n$ and task-specific termination head $g_n$. During execution, $g_n$ produces completion score $\beta_n(t)$; when $\beta_n(t)>\delta$, the controller activates the adapter for the next subtask.
}
\label{fig:teaser}
\end{figure*}

Existing approaches address this by explicitly modeling compositionality with the help of different data modalities,
such as language \cite{zhao2025cotvlavisualchainofthoughtreasoning} and images \cite{wang2026vlathinkerboostingvisionlanguageactionmodels}, actions \cite{zhong2026acotvlaactionchainofthoughtvisionlanguageaction}, or within the program space itself \cite{curtis2025proc3s,gao2025vlaosstructuringdissectingplanning}. Whether they structure how a policy is prompted or controlled~\cite{zawalski2025ecot, chen2026steerablevisionlanguageactionpoliciesembodied}, or decompose behaviors into simpler actions~\cite{bu2025univlalearningacttaskcentric, chen2025villaxenhancinglatentaction}, these approaches primarily structure composition in language, action, reasoning, or program space rather than adapting the policy weights themselves.
We argue instead that long-horizon composition should not rely solely on a monolithic policy to implicitly select the right parameters during its forward pass. It should also be addressed in the policy weight space, where task-specific semantics are explicitly encoded.
Furthermore, existing literature supports that systematic generalization through a compositional structure is most effective when known primitives can be recombined rather than relearned from scratch \cite{lake2023human, lake2016buildingmachineslearnthink, chollet2019measureintelligence, bianchi2026wizard}.
Inspired by this intuition, we introduce \modelname, a Skill-level Parameter Encoding framework for Long-horizon Learning. Fig.~\ref{fig:teaser} shows the central mechanism of \modelname.
For each subtask, corresponding to one manipulation skill in the ordered composition, we refer to its language instruction and short visual demonstration as \emph{task evidence}.
From this evidence, \modelname's meta-network generates a subtask-specific LoRA update that we call a \emph{skill-level adapter}. During execution, only one skill-level adapter is active at a time; a learned \emph{termination head} predicts a completion score and triggers the switch to the next adapter.
By training on task-specific adapters rather than composed-task demonstrations, \modelname reuses the knowledge already captured by expert policies while learning how this knowledge can be invoked across longer task sequences. This process is not tied to a single dataset or environment: we show adapters can come from heterogeneous sources \cite{khazatsky2025droidlargescaleinthewildrobot, liu2023liberobenchmarkingknowledgetransfer, mees2022calvinbenchmarklanguageconditionedpolicy} and can be jointly used to train the meta-network, making weight-space composition a practical mechanism for aggregating knowledge across different sources.

We evaluate \modelname on long-horizon simulated and real-world manipulation tasks while excluding the evaluated compositions from meta-training.
At deployment, \modelname conditions on target-task evidence but does not use target-task action labels or gradient-based optimization.
To summarize, our three contributions are:
\begin{itemize}
    \item We introduce \modelname, a sequential weight-space adaptation framework that generates skill-level low-rank adapters for a frozen VLA model and dynamically composes them during long-horizon execution.
    \item We train \modelname on single-task expert adapters, without composed-task demonstrations, and show that its meta-network can aggregate supervision from heterogeneous expert sources across different datasets and environments. We evaluate whether this supervision is sufficient to support compositional execution on held-out long-horizon tasks without target-task action supervision or gradient-based adaptation.
    \item We benchmark \modelname across CALVIN, LIBERO-10, Isaac Sim, and a real-world Franka Emika Panda setup, showing that skill-level adapter composition improves long-horizon execution over monolithic and non-compositional baselines.
\end{itemize}

%% file: sections/2_literature.tex
\section{Related Work}
\label{sec:related}

We review prior work along four axes: compositional and long-horizon robotic manipulation, generalist VLAs,
parameter-efficient and weight-space adaptation, and planning-based decomposition for long-horizon control.

\noindent \textbf{Compositional and long-horizon robotic manipulation.}
Compositional manipulation benchmarks evaluate whether policies
can combine existing skills in unseen scenarios.
CALVIN \cite{mees2022calvinbenchmarklanguageconditionedpolicy} introduces a standard long-horizon protocol in which policies are trained on environments A, B and C, and evaluated on the held-out D environment. LIBERO \cite{liu2023liberobenchmarkingknowledgetransfer} broadens it by
exposing the difficulty of lifelong learning and multitask transfer.
ClevrSkills \cite{haresh2024clevrskillscompositionallanguagevisual} 
isolates compositional language and visual reasoning to show how strong pretrained models still struggle to recombine learned skills. 
More recently, RoboCerebra \cite{han2025robocerebralargescalebenchmarklonghorizon}
extends long-horizon evaluation on LIBERO-style tasks and proposes a hierarchical framework for complex manipulation scenarios.
\modelname is complementary to hierarchical decomposition: it assumes an ordered subtask sequence and studies how the resulting skills can be instantiated and composed directly in weight space.

\noindent \textbf{Generalist vision-language-action policies.}
Large pretrained robot policies improve breadth and transfer by pooling experience from diverse robot datasets and embodiments. RT-1 \cite{brohan2023rt1}, RT-2 \cite{zitkovich2023rt2}, PaLM-E \cite{driess2023palme}, Octo \cite{ghosh2024octo}, OpenVLA \cite{kim2025openvla}, Flower \cite{reuss2025flowerdemocratizinggeneralistrobot}, $\pi_0$ and $\pi_{0.5}$ \cite{black2026pi0visionlanguageactionflowmodel, intelligence2025pi05visionlanguageactionmodelopenworld}, and GR00T \cite{nvidia2025gr00tn1openfoundation} show that large-scale pretraining can produce reusable vision-language-action backbones.
However, these monolithic policies rely on a single parameter set to select, sequence, and execute the skills required by a long-horizon task. In contrast, \modelname freezes the VLA backbone and uses generated skill-level adapters to specialize the policy for each subtask, ensuring both execution fidelity and broad transfer.

\noindent \textbf{Parameter-efficient and weight-space adaptation.}
Parameter-efficient adaptation provides a practical interface to specialize large policies without updating all model parameters. 
Methods such as LoRA \cite{hu2021loralowrankadaptationlarge} do so by learning low-rank updates, while structured adaptation methods steer policy behaviors through learned latent prompts that span a representation of the environment \cite{park2024foundationpolicieshilbertrepresentations}.
Prompt-to-weights generation further explores whether task evidence can be mapped directly to model updates \cite{liang2025draganddropllmszeroshotprompttoweights}.
These approaches, however, do not directly address long-horizon robotic composition. WIZARD \cite{bianchi2026wizard} introduces a meta-learning approach to generate LoRA parameters from task evidence, but struggles on compositional long-horizon tasks. \modelname addresses this gap by generating one skill-level adapter per subtask and composing them during a long-horizon rollout, treating the sequence of adapters itself as the compositional object.

\noindent \textbf{Planning, reasoning, and program-level decomposition.}
A complementary line of work tackles long-horizon composition by decomposing tasks into explicit reasoning steps, plans, or executable programs. Embodied Chain-of-Thought reasoning \cite{zawalski2025ecot} augments VLAs with grounded intermediate reasoning, SayCan \cite{ichter2023saycan}, Inner Monologue \cite{huang2023innermonologue}, and Trust the PRoC3S \cite{curtis2025proc3s} use language models for planning or feedback-based reasoning. Code as Policies \cite{huang2023codeaspolicies} and ShowTell \cite{murray2025showtell} synthesize code for robot programs from language and visual demonstrations. SteerablePolicies \cite{chen2026steerablevisionlanguageactionpoliciesembodied} improves instruction following by decomposing and enriching manipulation instructions. Instead of structuring behavior in a single domain space, \modelname targets a higher level of abstraction: it directly composes policy updates, allowing task-specific expertise to be activated directly in weight space during execution.
This makes \modelname complementary to explicit planners: the decomposition determines what to execute, while \modelname specializes how each subtask is executed in weight space.

%% file: sections/3_method.tex
\section{The SPELL framework}
\label{sec:method}
This section describes \modelname, a sequential weight-space adaptation framework for long-horizon compositional manipulation. Sec.~\ref{sec:problem_setup} formalizes the task setup for generating task-specific LoRA, while Sec.~\ref{sec:our_method} extends this formulation from single-task adaptation to sequential execution.

\subsection{Problem Setup and Notation}
\label{sec:problem_setup}

We study compositional long-horizon manipulation, where a task $\tau$ is defined as a sequence of $N$ subtasks 
\begin{equation}
\tau = (\tau_1,\ldots,\tau_N), \qquad \tau_n \in \mathcal{T},
\label{eq:comp_task}
\end{equation} 
where $\mathcal{T}$ denotes the set of available manipulation skills. Each subtask $\tau_n$ is described by a language instruction $p_n$ and by a short visual demonstration $v_n$. A policy rollout is successful only if all subtasks are completed in the prescribed order. If $S_n$ denotes successful completion of the $n$-th subtask, then
\begin{equation}
\mathrm{Success}(\tau)
=
\Pr\!\left(\bigcap_{n=1}^{N} S_n\right)
=
\prod_{n=1}^{N}\Pr\!\left(S_n \mid S_{<n}\right).
\label{eq:success_rate}
\end{equation}
Equation~\eqref{eq:success_rate} is the probability chain rule and does not impose conditional independence or a Markov assumption: each $S_n$ is evaluated from the state produced by the preceding subtasks, so inter-stage dependencies remain part of the rollout.
This formulation specifically highlights how stage-wise errors compound across longer horizons. Moreover, given $|\mathcal{T}| = M$, then the number of possible length-$N$ compositions grows as $M^N$, making it very difficult to learn a monolithic policy that truly excels in every task sequence.
\modelname addresses this limitation by lifting weight-space adaptation from single-task inference to sequential subtask composition. Instead of generating one low-rank adapter $\Delta{\widetilde{W}}$ for the full long-horizon task $\tau$, \modelname generates a sequence of subtask-specific adapters $\Delta{\widetilde{W}}_1,\dots,\Delta{\widetilde{W}}_N$, one per subtask, and executes them with an adaptive switching mechanism.

\noindent \textbf{Weight-space meta-learning preliminaries.}
We build on the weight-space meta-learning formulation introduced in~\cite{bianchi2026wizard}.
Let $G_W$ denote a pretrained Vision-Language-Action policy with frozen backbone parameters $W$.
For a task $\tau_i \in \mathcal{T}$, adapting the backbone with LoRA yields a task-specific expert update $\Delta W_i$.
We denote by $S_i$ the corresponding layer-wise scale statistics of
$\Delta W_i$, which are used as supervision for the scale-aware generator.
Each task is represented by a language instruction $p_i$ and a short visual demonstration $v_i$.
A frozen multimodal encoder maps this task evidence to
\begin{equation}
    z_i = G^{\mathrm{enc}}(p_i, v_i).
\end{equation}
As in~\citep{bianchi2026wizard}, we use a scale-aware meta-network $f_{\phi}$ to generate a task-specific LoRA update $\Delta \widetilde{W}_i$ together with its predicted layer-wise scale statistics $\widetilde{S}_i$ from the task embedding $z_i$.
The generated update is then rescaled before injection into the frozen backbone:
\begin{equation}
\Delta \widetilde{W}_i
\leftarrow
\widetilde{S}_i
\frac{\Delta \widetilde{W}_i}{\|\Delta \widetilde{W}_i\|}.
\label{eq:rescale}
\end{equation}
This preserves the numerical stability of weight-space generation for heterogeneous VLA modules.

\subsection{Sequential Skill-level Adapter Composition}
\label{sec:our_method}
This section describes the \modelname\ framework summarized in Fig.~\ref{fig:teaser}.
We first introduce how \modelname\ generates a specialized LoRA adapter for each subtask of a long-horizon task and executes the resulting experts sequentially.
We then introduce the mechanism used to detect subtask completion and switch from the current expert to the next one.

\noindent \textbf{Subtask Adapter Generation.}
Given a composed task $\tau$, defined as an ordered list of subtasks in
Eq.~\eqref{eq:comp_task}, we define the corresponding sequence of subtask
evidences as
\begin{equation}
    \big\{(p_n, v_n)\big\}_{n=1}^{N}.
\end{equation}
For each subtask, we extract the corresponding task embedding
\begin{equation}
    z_n = G^{\mathrm{enc}}(p_n, v_n),
\end{equation}
from which we obtain a task-specific LoRA update
$\Delta \widetilde{W}_n$ and its predicted layer-wise scale statistics
$\widetilde{S}_n$.
The generated update is then rescaled using Eq.~\eqref{eq:rescale}.
At time $t$, $o_t$ denotes the current visual observation, $s_t$ the robot proprioceptive state, and $a_t$ the control action. The $n$-th generated adapter is injected into the frozen VLA backbone to instantiate the $n$-th expert policy
\begin{equation}
\pi_n(a_t \mid o_t, s_t, p_n)
:=
G_{W + \Delta \widetilde{W}_n}(a_t \mid o_t, s_t, p_n).
\end{equation}
Rather than storing all experts simultaneously, we generate them on demand and swap only the active LoRA parameters.
This requires storing only the frozen backbone and the currently active adapter, rather than all subtask experts simultaneously.

\noindent \textbf{Termination-based expert switching.}
Sequential execution requires a mechanism for determining when the active subtask has been completed and the next expert should be activated.
We therefore associate each task-specific expert with a lightweight \emph{termination head}.
For the source experts, its parameters are denoted by $\theta_n$, while at inference time the meta-network generates the corresponding parameters $\tilde{\theta}_n$.
The generated termination head estimates the progress of the current subtask from the current visual observation, proprioceptive state, and task embedding:
\begin{equation}
    \beta_n(t)
    =
    \sigma\!\left(g_n(o_t,s_t,z_n;\tilde{\theta}_n)\right)
    \in [0,1].
\end{equation}

For each source expert, the termination head is trained from the same
demonstrations used to train the corresponding LoRA adapter, while keeping
the VLA backbone and task-specific LoRA parameters frozen.
We denote its predicted progress by
\begin{equation}
    \beta_n^{\mathrm{src}}(t)
    =
    \sigma\!\left(g_n(o_t,s_t,z_n;\theta_n)\right).
\end{equation}
For an episode $k$ of length $T^{(k)}$, we use normalized trajectory progress
$y_t^{(k)}=t/T^{(k)}$ as dense supervision and minimize
\begin{equation}
    \mathcal{L}_{\mathrm{term}}
    =
    \frac{1}{|\mathcal{B}|}
    \sum_{(k,t)\in\mathcal{B}}
    \left|
    \beta_{n,\mathrm{src}}^{(k)}(t)-\frac{t}{T^{(k)}}
    \right|.
\end{equation}
This provides a smooth progress signal rather than supervision only at the terminal frame.

The resulting termination-head parameters are treated as part of the
task-specific expert representation. Accordingly, the meta-network jointly
generates the LoRA update, its layer-wise scale statistics, and the
termination-head parameters from the same task embedding:
\begin{equation}
    (\Delta \widetilde{W}_n, \widetilde{S}_n, \tilde{\theta}_n)
    = f_\phi(z_n).
\label{eq:spell_generation}
\end{equation}
During meta-network training, the source termination-head parameters
$\theta_n$ are treated as an additional weight-space supervision target,
alongside the source LoRA parameters $\Delta W_n$ and their layer-wise
scale statistics $S_n$. Thus, $f_\phi$ learns to predict
$\tilde{\theta}_n$ jointly with $\Delta\widetilde{W}_n$ and
$\widetilde{S}_n$ from the same task evidence.

At inference time, \modelname processes the subtasks sequentially.
For the current subtask $\tau_n$, the meta-network generates
$(\Delta\widetilde{W}_n,\widetilde{S}_n,\tilde{\theta}_n)$ from $z_n$.
The generated adapter is injected into the frozen backbone to instantiate
$\pi_n$, which is executed while the corresponding termination head
continuously evaluates $\beta_n(t)$.
When $\beta_n(t)>\delta$, where $\delta$ is a completion threshold,
the current subtask is considered complete and the controller proceeds to
$\tau_{n+1}$, repeating the same generation--execution--termination cycle.
We additionally use a conservative timeout fallback for robustness.

\noindent \textbf{Bridge reset.}
When enabled for a given evaluation setting, we apply a \emph{bridge reset} that returns the
manipulator to a neutral configuration while preserving the scene state.
This reduces the distribution shift between the terminal state of the
completed subtask and the initial state expected by the next expert,
without undoing the physical consequences of previous subtasks.
The specific reset strategy adopted for each evaluation setting is
described in the corresponding experimental section.

%% file: sections/4_experiments.tex
\section{Experiments}
\label{sec:expr}

Sec.~\ref{sec:experimental_setup} describes the data used to train \modelname, the training protocols, and the evaluation benchmarks used to measure long-horizon compositional generalization under held-out task compositions.
In Sec.~\ref{sec:calvin_exps} we evaluate on CALVIN~\cite{mees2022calvinbenchmarklanguageconditionedpolicy},
and in Sec.~\ref{sec:libero_exps} on LIBERO-10~\cite{liu2023liberobenchmarkingknowledgetransfer}.
Sec.~\ref{sec:isaac_exps} evaluates state-dependent compositions in Isaac Sim~\cite{nvidia2025isaaclabgpuacceleratedsimulation}
using DROID-style policies~\cite{khazatsky2025droidlargescaleinthewildrobot},
while Sec.~\ref{sec:real_world_exps} evaluates real-world deployment on a Franka Emika Panda.
Finally, in Sec.~\ref{sec:ablations}, we complement the main results with controlled analyses of the design choices and the volume and diversity of expert supervision used to train \modelname.

\subsection{Experimental Setup}
\label{sec:experimental_setup}

\noindent\textbf{Expert training.}
We independently train one single-task expert per source task using the same imitation-learning objective as the base VLA. Experts use rank-16 LoRA ($\alpha=16$) on all linear modules of the frozen $\pi_{0.5}$ backbone, implemented with PEFT, zero dropout, and Gaussian initialization. For CALVIN, we first adapt $\pi_{0.5}$ to the ABC source distribution and then train task-specific experts from this source-domain initialization; for LIBERO and CALVIN, we use an action horizon of 10. Real-world experts are trained analogously from DROID-style and single-task target-setup data.
Task evidence is extracted from the frozen $\pi_{0.5}$ multimodal prefix encoder without action labels and mean-pooled over the demonstration; during meta-training, each expert adapter is paired with a compatible episode embedding from the same source task.
The available expert pool contains 10 LIBERO-Goal, 10 Object, 10 Spatial, 76 LIBERO-90, 343 CALVIN-ABC, and 149 DROID experts. DROID experts are used for the real-world setting and separately in the cross-domain scaling analysis below.

\noindent\textbf{Termination-head training.}
Each source expert is paired with a residual MLP termination head $g_n$, conditioned on the 2048-D task embedding $z_n$, with LayerNorm, a two-layer GELU MLP with dropout, a second LayerNorm, and a scalar sigmoid output. It is trained after the corresponding single-task expert, with the VLA backbone and LoRA adapter frozen, using AdamW with learning rate $10^{-4}$, weight decay $10^{-5}$, batch size 256, and 1000 optimization steps. Its trained parameters are included in the expert target and generated jointly with the LoRA update by the meta-network. We set the completion threshold to $\delta=0.9$ based on validation rollouts and use a conservative timeout fallback at inference.

\noindent\textbf{Training and evaluation split.}
Training and evaluation tasks are kept strictly separated.
For LIBERO-10, experts and the meta-network are trained on LIBERO-Spatial, LIBERO-Object, LIBERO-Goal, LIBERO-90, and CALVIN-ABC, with LIBERO-10 held out entirely.
For CALVIN-D, environment D is similarly excluded from training.
Likewise, the DROID data used for real-world expert supervision do not contain the composed sequences used for evaluation.
Thus, the meta-network never sees any of the composed sequences used for evaluation.

\noindent\textbf{Meta-network training.}
Given the pool of trained source experts, we train the meta-network from task evidence consisting of a short visual demonstration and a language instruction.
The meta-network is a 5-layer 3D CNN with kernel size 7, hidden dimension 512, and 2048-D input. It predicts the LoRA update, per-layer scale statistics used by the scale-aware normalization in Eq.~\ref{eq:rescale}, and the parameters of the task-specific termination head.
We train it for 5000 steps with AdamW (learning rate $10^{-4}$, weight decay $0.1$), batch size 6 (effective 24), and gradient clipping at $1.0$.

\noindent\textbf{Evaluation protocol.}
At test time, \modelname receives a target-task episode as task evidence and discards its action labels; no target-task parameter update, finetuning, or gradient step is performed. When a comparison method performs target-task gradient adaptation, such as adapted $\pi_{0.5}$, it is allowed to use the episode's action labels according to its protocol, while published benchmark baselines follow their standard evaluation settings. All methods that require an explicit decomposition use the same ordered subtask sequence.
Benchmark-specific execution protocols are detailed in the corresponding evaluation sections.

\subsection{CALVIN: Long-Horizon Skill Composition}
\label{sec:calvin_exps}

\noindent\textbf{Evaluation setup and baselines.}
CALVIN is a language-conditioned long-horizon manipulation benchmark with a 7-DoF Franka Emika Panda and four structurally related environments, A--D, sharing task semantics while varying scene layouts and object configurations. We follow the standard $ABC \rightarrow D$ protocol: environments A, B, and C are used for source experts and meta-training, while D is held out for evaluation.

At test time, each rollout contains a closed-loop sequence of up to five language-conditioned instructions. The scene state is preserved across subtasks and no bridge reset is applied. We report $SR_n$, the fraction of rollouts completing at least $n$ consecutive instructions, and Avg., the average number of completed instructions per sequence. Thus, the evaluation directly measures how execution errors accumulate with horizon.
We compare \modelname against recent CALVIN baselines, such as~\cite{reuss2024efficientdiffusiontransformerpolicies, tian2024predictiveinversedynamicsmodels, hu2025videopredictionpolicygeneralist, bu2025univlalearningacttaskcentric, reuss2025flowerdemocratizinggeneralistrobot}, the base $\pi_{0.5}$ policy, and compositional adapters  generated with WIZARD~\cite{bianchi2026wizard}. 

\noindent\textbf{Results.}
Table~\ref{tab:calvin_results} shows that \modelname achieves the highest average sequence length of $4.58$, compared with $4.53$ for FLOWER, $4.42$ for $\pi_{0.5}$, and $4.11$ for WIZARD. While remaining competitive at short horizons, \modelname achieves the best performance from $SR_3$ onward, reaching $0.92$, $0.89$, and $0.86$ on $SR_3$, $SR_4$, and $SR_5$, respectively. The advantage increases with the execution horizon: at $SR_5$, \modelname improves over FLOWER and $\pi_{0.5}$ by $8$ and $9$ percentage points, respectively, and over WIZARD by $20$ points. These results support the benefit of adapting the policy with subtask-specific experts during sequential execution, particularly as longer skill sequences are composed.

\input{tables/CALVIN_main_results}

\subsection{LIBERO-10: Held-Out Compositional Evaluation}
\label{sec:libero_exps}

\input{tables/LIBERO_10_main_results}

\noindent\textbf{Evaluation setup and baselines.}
LIBERO is a language-conditioned tabletop manipulation benchmark built on Robosuite. Its source suites isolate complementary forms of variation: LIBERO-Spatial, LIBERO-Object, and LIBERO-Goal each contain 10 tasks emphasizing spatial relations, object identities, and task goals, while LIBERO-90 provides a broader collection of short-horizon skills. We use these suites only as sources of single-task experts and meta-network supervision; LIBERO-10 is held out for compositional evaluation.
We compare against miniVLA-VQ~\cite{belkhale2024minivla}, InSpire~\cite{zhang2025inspirevisionlanguageactionmodelsintrinsic}, $\pi_{0.5}$, and WIZARD~\cite{bianchi2026wizard}.
At each LIBERO-10 subtask transition, we apply a \emph{bridge reset}: the manipulator is returned to its neutral configuration while the scene state is preserved. This reduces inter-subtask distribution shift without undoing the physical effects of the completed subtask.

Success is measured through simulator predicates that verify the symbolic goal state, e.g., placing an object in a container, moving it relative to another object, manipulating articulated objects, or toggling appliances. We report success for the composed LIBERO-10 tasks together with macro-averages over the first eight tasks and the full ten-task suite. T6 contains only one evaluable stage and is retained in the reported averages for consistency.

\noindent\textbf{Results.}
Table~\ref{tab:libero_results} shows that \modelname substantially improves compositional performance on LIBERO-10, reaching $\mathrm{Avg.}(8)=0.62$ and $\mathrm{Avg.}(10)=0.50$. This represents a large improvement over the strongest comparison method, InSpire, which reaches $0.29$ and $0.25$, respectively, while $\pi_{0.5}$ obtains $0.24/0.20$ and WIZARD $0.02/0.01$. The largest gains appear on T1, T2, and T8, where \modelname reaches $0.86$, $1.00$, and $0.92$, compared with the strongest baseline results of $0.54$, $0.56$, and $0.26$, respectively. Overall, these results show that sequential generation and execution of subtask-specific adapters substantially improves held-out compositional generalization over both direct VLA adaptation and single-adapter weight-space generation.

\subsection{State-Dependent Composition in Isaac Sim}
\label{sec:isaac_exps}

\noindent\textbf{Evaluation setup.}
We evaluate \modelname in Isaac Sim~\cite{nvidia2025isaaclabgpuacceleratedsimulation} using DROID-style policies~\cite{khazatsky2025droidlargescaleinthewildrobot}. We consider three held-out four-stage compositions: (i) relocate a bowl and place a cube inside it; (ii) relocate a mug and place a can inside it; and (iii) relocate a bin and place a banana inside it. Each sequence is explicitly state dependent: the receptacle is first relocated, and the subsequent manipulation must therefore operate on the scene state produced by the preceding stages. Each method is evaluated over 50 episodes per composition using matched initial scene configurations, and a rollout is considered successful only if all four stages are completed in order. Isaac Sim trajectories are used only for evaluation and are not included in meta-training.

\noindent\textbf{Results.}
Fig.~\ref{fig:isaac_results} shows that \modelname improves full-sequence success across the state-dependent Isaac Sim evaluation, reaching a mean success rate of $0.41$ compared with $0.29$ for adapted $\pi_{0.5}$. This corresponds to a $12$ percentage-point absolute improvement despite requiring all four stages to succeed consecutively. Since the second half of each composition must operate on the scene produced by the first, these results indicate that sequentially generated adapters can remain effective under inter-stage state changes rather than only when subtasks can be treated independently.
The improvement is consistent across all three compositions: $0.62$ vs.\ $0.48$ for bowl--cube, $0.24$ vs.\ $0.18$ for mug--can, and $0.36$ vs.\ $0.20$ for bin--banana.

\input{tables/isaac_sim_results}

\subsection{Real-World Evaluation}
\label{sec:real_world_exps}

\noindent\textbf{Franka Emika Panda.}
We evaluate on a 7-DoF Franka Emika Panda with a parallel gripper in a DROID-style tabletop setup. Visual observations are captured from three Intel RealSense cameras: a wrist-mounted D415 and two external D405 cameras. Policy outputs are executed through a joint-space impedance controller at $1$\,kHz, while high-level policy inference runs at $15$\,Hz. Both the adapted $\pi_{0.5}$ baseline and \modelname start from the same real-domain initialization: a $\pi_{0.5}$ checkpoint pretrained on DROID and adapted to the target embodiment using teleoperated single-task data.

We evaluate three manipulation skills (picking an apple, a banana, and a cup) composed in different orders into sequences of increasing horizon. Over 30 physical episodes, we report $SR_n$, the fraction of sequences completing at least the first $n$ consecutive skills, and Mean SR across the evaluated horizons. Fig.~\ref{fig:franka_results} shows that \modelname consistently outperforms adapted $\pi_{0.5}$ as skills are composed: $SR_1/SR_2/SR_3$ increases from $0.57/0.27/0.10$ to $0.87/0.53/0.27$, while Mean SR improves from $0.31$ to $0.56$. In relative terms, the advantage grows with sequence horizon, with \modelname nearly doubling $SR_2$ and reaching $2.7\times$ the baseline success at $SR_3$. These results show that dynamically generating and switching subtask-specific adapters improves execution reliability across real-world skill compositions.

\input{tables/real_world_results}

\subsection{Controlled Analysis}
\label{sec:ablations}

Fig.~\ref{fig:libero_ablations} shows the controlled analyses of \modelname. Panel (a) shows that richer task evidence improves performance, with video conditioning outperforming language-only and single-frame conditioning. Panel (b) shows that the bridge reset improves execution while preserving \modelname's advantage under matched reset conditions. Panel (c) shows that termination-based switching approaches oracle switching and outperforms a fixed-duration schedule. Panel (d) shows that neither a single composition-level WIZARD adapter nor nearest-neighbor retrieval from the expert pool explains \modelname's performance; the remaining gap to oracle experts instead indicates headroom in adapter generation. Finally, panel (e) shows that CALVIN-D performance improves as increasingly diverse related experts are added to the expert pool, while adding DROID experts produces no measurable cross-domain interference.
Quantitatively, video conditioning reaches $0.50$ average success compared with $0.46$ for a single frame and $0.34$ for language alone, while learned termination reaches $0.50$, compared with $0.54$ for oracle switching and $0.38$ for the fixed-duration schedule.

\input{tables/LIBERO_10_ablations}

The reset comparison aggregates 1,500 rollouts; under matched reset conditions, \modelname significantly outperforms both $\pi_{0.5}$ and WIZARD ($p<0.001$, paired permutation test over matched task--seed averages). The termination head achieves precision/recall $0.94/0.91$ and crosses its threshold on only $6\%$ of failed rollouts. Timestep, elapsed time, and episode duration are not provided to the detector at inference.

On an A6000, adapter generation requires $0.85$\,s ($0.78$\,s encoding and $0.07$\,s generation) and adapter switching $1.23$\,s. On LIBERO-10, the $4.50$\,s bridge reset runs in parallel and dominates transition latency. On-demand execution stores only the frozen backbone and active adapter, requiring $O(|W|+|\Delta W|)$ memory independently of sequence horizon.

%% file: tables/CALVIN_main_results.tex
\begin{table}[t]
\centering

\setlength{\tabcolsep}{3pt}
\renewcommand{\arraystretch}{0.95}

\resizebox{\columnwidth}{!}{%
\begin{tabular}{@{}lcccccc@{}}
\toprule
\textbf{Method}
& $\mathbf{SR_1}$
& $\mathbf{SR_2}$
& $\mathbf{SR_3}$
& $\mathbf{SR_4}$
& $\mathbf{SR_5}$
& \textbf{Avg.} \\
\midrule

MODE
& 0.96 & 0.91 & 0.81 & 0.72 & 0.64 & 4.01 \\

SeeR-Large
& 0.96 & 0.92 & 0.86 & 0.80 & 0.74 & 4.28 \\

VPP
& 0.96 & 0.92 & 0.86 & 0.81 & 0.75 & 4.29 \\

UniVLA
& \textbf{0.99} & \underline{0.95} & 0.89 & 0.83 & 0.75 & 4.41 \\

FLOWER
& \textbf{0.99}
& \textbf{0.96}
& \underline{0.91}
& \underline{0.85}
& \underline{0.78}
& \underline{4.53} \\

$\pi_{0.5}$
& 0.97
& 0.94
& 0.90
& 0.83
& 0.77
& 4.42 \\

WIZARD
& 0.96 & 0.92 & 0.83 & 0.74 & 0.66 & 4.11 \\

\rowcolor{gray!15}
\textbf{\modelname}
& \underline{0.98}
& 0.94
& \textbf{0.92}
& \textbf{0.89}
& \textbf{0.86}
& \textbf{4.58} \\

\bottomrule
\end{tabular}%
}

\caption{
    \textbf{Performance on CALVIN ($ABC \rightarrow D$).}
    $SR_n$ denotes success after completing $n$ consecutive
    instructions. Bold and underline indicate the best and
    second-best results, respectively.
}
\label{tab:calvin_results}

\end{table}

%% file: tables/LIBERO_10_main_results.tex
\begin{table*}[!t]
\centering
\footnotesize
\renewcommand{\arraystretch}{1.08}

\begin{tabular*}{\textwidth}{
    @{\extracolsep{\fill}}
    l
    cccccccccc
    | cc
    @{}
}
\toprule
\multirow{2}{*}{\textbf{Method}}
& \multicolumn{10}{c}{\textbf{Individual Task Success Rates}}
& \multicolumn{2}{c}{\textbf{Macro Averages}} \\
\cmidrule(lr){2-11}
\cmidrule(lr){12-13}

& \textbf{T1}
& \textbf{T2}
& \textbf{T3}
& \textbf{T4}
& \textbf{T5}
& \textbf{T6$^\dagger$}
& \textbf{T7}
& \textbf{T8}
& \textbf{T9}
& \textbf{T10}
& \textbf{Avg. (8)}
& \textbf{Avg. (10)} \\
\midrule

miniVLA-VQ
& 0.21 & 0.16 & 0.39 & \textbf{0.56}
& 0.01 & 0.32 & 0.01 & 0.15 & 0.15 & 0.00
& 0.23 & 0.20 \\

InSpire
& 0.54 & 0.14 & 0.34 & \textbf{0.56}
& 0.06 & 0.34 & 0.09 & 0.26 & \textbf{0.17} & 0.00
& 0.29 & 0.25 \\

$\pi_{0.5}$
& 0.19 & 0.56 & 0.26 & 0.08
& \textbf{0.36} & 0.22 & 0.17 & 0.11 & 0.00 & 0.00
& 0.24 & 0.20 \\

WIZARD
& 0.00 & 0.00 & 0.00 & 0.00
& 0.00 & 0.14 & 0.00 & 0.00 & 0.00 & 0.00
& 0.02 & 0.01 \\

\rowcolor{gray!15}
\textbf{\modelname}
& \textbf{0.86}
& \textbf{1.00}
& \textbf{0.50}
& 0.52
& \textbf{0.36}
& \textbf{0.42}
& \textbf{0.41}
& \textbf{0.92}
& 0.01
& 0.00
& \textbf{0.62}
& \textbf{0.50} \\

\bottomrule
\end{tabular*}

\caption{\textbf{Performance on LIBERO-10.}
Success rates for each task and macro-averages over the first eight
and all ten tasks. $^\dagger$T6 contains only one evaluable stage
and is retained for benchmark consistency.}
\label{tab:libero_results}
\end{table*}

%% file: tables/isaac_sim_results.tex
\begin{figure}[t]
\centering

\begin{tikzpicture}
\begin{axis}[
    width=0.82\columnwidth,
    height=2.55cm,
    scale only axis,
    ybar,
    bar width=8pt,
    ymin=0,
    ymax=0.8,
    xmin=0.45,
    xmax=3.55,
    xtick={1,2,3},
    xticklabels={
        {Bowl $\rightarrow$ cube},
        {Mug $\rightarrow$ can},
        {Bin $\rightarrow$ banana}
    },
    ytick={0,0.2,0.4,0.6,0.8},
    ylabel={Full-sequence success},
    axis x line*=bottom,
    axis y line*=left,
    tick align=outside,
    ylabel style={font=\scriptsize},
    xticklabel style={
        font=\tiny,
        align=center
    },
    yticklabel style={font=\scriptsize},
    legend columns=2,
    legend style={
        draw=none,
        font=\scriptsize,
        at={(0.5,1.03)},
        anchor=south,
        /tikz/every even column/.append style={
            column sep=8pt
        }
    }
]

\addplot[
    bar shift=-4.5pt,
    fill=picolor!35,
    draw=picolor,
    line width=0.8pt
]
coordinates {
    (1,0.48)
    (2,0.18)
    (3,0.20)
};
\addlegendentry{Adapted $\pi_{0.5}$}

\addplot[
    bar shift=4.5pt,
    fill=ourscolor!85,
    draw=ourscolor,
    line width=0.8pt
]
coordinates {
    (1,0.62)
    (2,0.24)
    (3,0.36)
};
\addlegendentry{\modelname}

\node[
    anchor=north east,
    font=\tiny,
    align=right,
    fill=white,
    fill opacity=0.90,
    text opacity=1,
    inner sep=1.5pt
] at (rel axis cs:0.98,0.97) {
    Mean $\pi_{0.5}=0.29$\\
    Mean \modelname$=0.41$
};

\end{axis}
\end{tikzpicture}

\caption{
\textbf{State-dependent composition in Isaac Sim.}
Full-sequence success on three held-out four-stage compositions in which later skills operate on scene states produced by preceding stages.
}
\label{fig:isaac_results}

\end{figure}
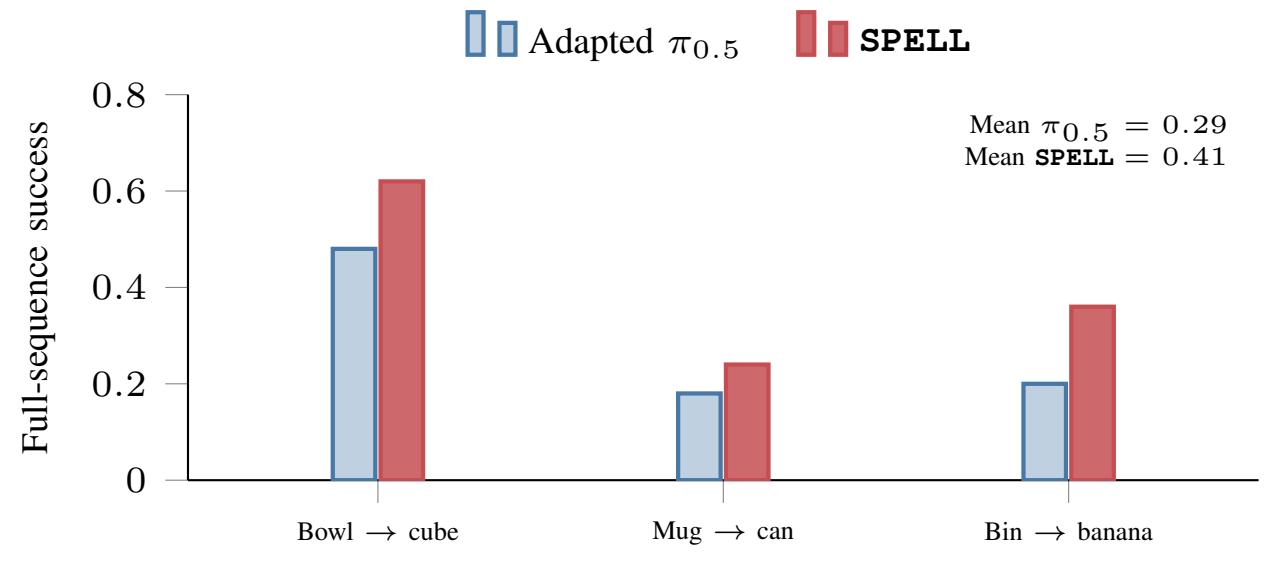

%% file: tables/real_world_results.tex
\providecolor{picolor}{RGB}{76,114,176}
\providecolor{ourscolor}{RGB}{196,78,82}

\begin{figure}[t]
\centering

\begin{tikzpicture}
\begin{axis}[
    width=\columnwidth,
    height=4cm,
    ymin=0,
    ymax=1.05,
    xmin=0.8,
    xmax=3.2,
    xtick={1,2,3},
    ytick={0,0.2,0.4,0.6,0.8,1.0},
    xlabel={Composition horizon $n$},
    ylabel={Success rate $SR_n$},
    axis x line*=bottom,
    axis y line*=left,
    tick align=outside,
    ylabel style={font=\tiny},
    xlabel style={font=\tiny},
    xticklabel style={font=\tiny},
    yticklabel style={font=\tiny},
    legend columns=2,
    legend style={
        draw=none,
        font=\tiny,
        at={(0.5,1.02)},
        anchor=south,
        /tikz/every even column/.append style={
            column sep=6pt
        }
    }
]

\addplot[
    draw=picolor,
    dashed,
    line width=1.0pt,
    mark=square,
    mark size=2.2pt,
    mark options={
        solid,
        draw=picolor,
        fill=white,
        line width=0.7pt
    }
]
coordinates {
    (1,0.57)
    (2,0.27)
    (3,0.10)
};
\addlegendentry{Adapted $\pi_{0.5}$}

\addplot[
    draw=ourscolor,
    solid,
    line width=1.15pt,
    mark=*,
    mark size=2.3pt,
    mark options={
        solid,
        draw=ourscolor,
        fill=ourscolor,
        line width=0.7pt
    }
]
coordinates {
    (1,0.87)
    (2,0.53)
    (3,0.27)
};
\addlegendentry{\modelname}

\node[
    anchor=north east,
    font=\tiny,
    align=right,
    fill=white,
    fill opacity=0.90,
    text opacity=1,
    inner sep=1.2pt
] at (rel axis cs:0.98,0.97) {
    Mean SR $\pi_{0.5}=0.31$\\
    Mean SR \modelname$=0.56$
};

\end{axis}
\end{tikzpicture}

\caption{
\textbf{Real-world long-horizon composition.}
Prefix success as increasingly many manipulation skills are executed consecutively on the Franka Emika Panda.
}
\label{fig:franka_results}

\end{figure}
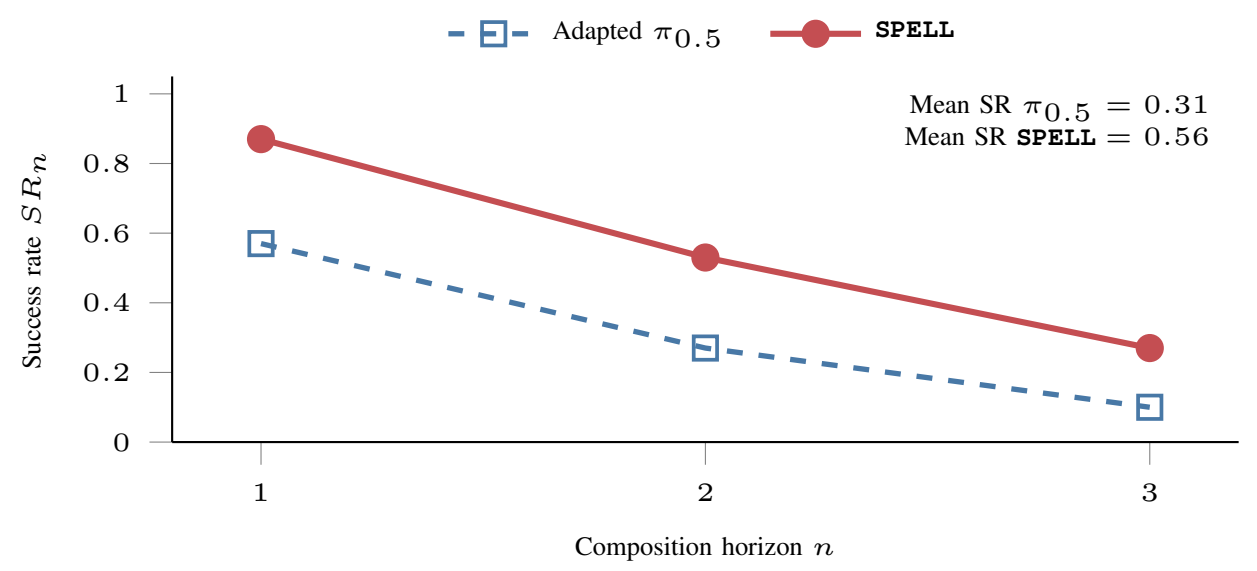

%% file: tables/LIBERO_10_ablations.tex
\providecolor{picolor}{RGB}{76,114,176}
\providecolor{nncolor}{RGB}{85,168,104}
\providecolor{oftcolor}{RGB}{221,132,82}
\providecolor{ourscolor}{RGB}{196,78,82}
\providecolor{oraclecolor}{RGB}{100,100,100}

\begin{figure*}[t]
\centering

\begin{tikzpicture}
\begin{groupplot}[
    group style={group size=3 by 1,horizontal sep=0.85cm},
    height=2.45cm,
    scale only axis,
    ymin=0,
    axis x line*=bottom,
    axis y line*=left,
    tick align=outside,
    ylabel style={font=\scriptsize},
    xlabel style={font=\scriptsize},
    xticklabel style={font=\tiny,align=center},
    yticklabel style={font=\scriptsize},
    title style={font=\scriptsize\bfseries,yshift=2pt}
]

\nextgroupplot[
    width=0.275\textwidth,
    title={(a) Conditioning evidence},
    ylabel={Avg. success},
    ymin=0,
    ymax=0.57,
    ytick={0,0.1,0.2,0.3,0.4,0.5},
    xmin=0.5,
    xmax=3.5,
    xtick={1,2,3},
    xticklabels={Language,Initial frame,Video}
]

\addplot[
    ybar,
    bar width=10pt,
    fill=ourscolor!35,
    draw=ourscolor,
    line width=0.8pt,
    error bars/.cd,
    y dir=both,
    y explicit,
    error bar style={draw=black}
]
coordinates {
    (1,0.34) +- (0,0.04)
};

\addplot[
    ybar,
    bar width=10pt,
    fill=ourscolor!65,
    draw=ourscolor,
    line width=0.8pt,
    error bars/.cd,
    y dir=both,
    y explicit,
    error bar style={draw=black}
]
coordinates {
    (2,0.46) +- (0,0.04)
};

\addplot[
    ybar,
    bar width=10pt,
    fill=ourscolor,
    draw=ourscolor,
    line width=0.8pt,
    error bars/.cd,
    y dir=both,
    y explicit,
    error bar style={draw=black}
]
coordinates {
    (3,0.50) +- (0,0.02)
};

\nextgroupplot[
    width=0.19\textwidth,
    title={(b) Bridge reset},
    ylabel={Avg. success},
    ymin=0,
    ymax=0.57,
    ytick={0,0.1,0.2,0.3,0.4,0.5},
    xmin=0.5,
    xmax=2.5,
    xtick={1,2},
    xticklabels={No reset,Bridge reset},
    legend to name=methodlegend,
    legend columns=5,
    legend style={
        draw=none,
        font=\scriptsize,
        /tikz/every even column/.append style={column sep=7pt},
        legend image post style={scale=0.8}
    }
]

\addlegendimage{area legend,fill=picolor,draw=black}
\addlegendentry{$\pi_{0.5}$}

\addlegendimage{area legend,fill=oftcolor,draw=black}
\addlegendentry{WIZARD}

\addlegendimage{area legend,fill=nncolor,draw=black}
\addlegendentry{NN retrieval}

\addlegendimage{area legend,fill=ourscolor,draw=black}
\addlegendentry{\modelname}

\addlegendimage{area legend,fill=oraclecolor,draw=black}
\addlegendentry{Oracle}

\addplot[
    ybar,
    bar width=7pt,
    bar shift=-8pt,
    fill=picolor,
    draw=black,
    error bars/.cd,
    y dir=both,
    y explicit,
    error bar style={draw=black,xshift=-8pt}
]
coordinates {
    (1,0.08) +- (0,0.01)
    (2,0.20) +- (0,0.02)
};

\addplot[
    ybar,
    bar width=7pt,
    bar shift=0pt,
    fill=oftcolor,
    draw=black,
    error bars/.cd,
    y dir=both,
    y explicit,
    error bar style={draw=black,xshift=0pt}
]
coordinates {
    (1,0.01) +- (0,0.01)
    (2,0.01) +- (0,0.01)
};

\addplot[
    ybar,
    bar width=7pt,
    bar shift=8pt,
    fill=ourscolor,
    draw=black,
    error bars/.cd,
    y dir=both,
    y explicit,
    error bar style={draw=black,xshift=8pt}
]
coordinates {
    (1,0.13) +- (0,0.02)
    (2,0.50) +- (0,0.02)
};

\nextgroupplot[
    width=0.275\textwidth,
    title={(c) Switching rule},
    ylabel={Avg. success},
    ymin=0,
    ymax=0.60,
    ytick={0,0.1,0.2,0.3,0.4,0.5,0.6},
    xmin=0.5,
    xmax=3.5,
    xtick={1,2,3},
    xticklabels={Fixed duration,Learned,Oracle}
]

\addplot[
    ybar,
    bar width=13pt,
    fill=ourscolor!35,
    draw=ourscolor,
    line width=0.8pt
]
coordinates {
    (1,0.38)
};

\addplot[
    ybar,
    bar width=13pt,
    fill=ourscolor,
    draw=ourscolor,
    line width=0.8pt
]
coordinates {
    (2,0.50)
};

\addplot[
    ybar,
    bar width=13pt,
    fill=oraclecolor,
    draw=black,
    line width=0.8pt
]
coordinates {
    (3,0.54)
};

\end{groupplot}
\end{tikzpicture}

\pgfplotslegendfromname{methodlegend}

\begin{tikzpicture}
\begin{groupplot}[
    group style={group size=2 by 1,horizontal sep=1.25cm},
    height=2.55cm,
    scale only axis,
    axis x line*=bottom,
    axis y line*=left,
    tick align=outside,
    ylabel style={font=\scriptsize},
    xlabel style={font=\scriptsize},
    xticklabel style={font=\scriptsize,align=center},
    yticklabel style={font=\scriptsize},
    title style={font=\scriptsize\bfseries,yshift=2pt}
]

\nextgroupplot[
    width=0.34\textwidth,
    title={(d) Adapter source},
    ylabel={Avg. success},
    ymin=0,
    ymax=0.88,
    ytick={0,0.2,0.4,0.6,0.8},
    xmin=0.5,
    xmax=4.5,
    xtick={1,2,3,4},
    xticklabels={Fixed WIZARD,NN retrieval,\modelname,Oracle}
]

\addplot[
    ybar,
    bar width=12pt,
    fill=oftcolor,
    draw=black,
    error bars/.cd,
    y dir=both,
    y explicit,
    error bar style={draw=black}
]
coordinates {
    (1,0.01) +- (0,0.01)
};

\addplot[
    ybar,
    bar width=12pt,
    fill=nncolor,
    draw=black,
    error bars/.cd,
    y dir=both,
    y explicit,
    error bar style={draw=black}
]
coordinates {
    (2,0.04) +- (0,0.01)
};

\addplot[
    ybar,
    bar width=12pt,
    fill=ourscolor,
    draw=black,
    error bars/.cd,
    y dir=both,
    y explicit,
    error bar style={draw=black}
]
coordinates {
    (3,0.50) +- (0,0.02)
};

\addplot[
    ybar,
    bar width=12pt,
    fill=oraclecolor,
    draw=black,
    error bars/.cd,
    y dir=both,
    y explicit,
    error bar style={draw=black}
]
coordinates {
    (4,0.80) +- (0,0.02)
};

\nextgroupplot[
    width=0.41\textwidth,
    title={(e) Expert-pool scaling},
    ylabel={$\Delta$ Avg. vs.\ ABC},
    xlabel={Cumulative expert sources},
    xmin=0.7,
    xmax=6.3,
    ymin=-0.01,
    ymax=0.23,
    xtick={1,2,3,4,5,6},
    xticklabels={ABC,+Goal,+Object,+Spatial,+LIBERO-90,+DROID},
    xticklabel style={font=\tiny,rotate=27,anchor=east},
    ytick={0,0.05,0.10,0.15,0.20},
    yticklabels={0,.05,.10,.15,.20},
    grid=major,
    grid style={line width=0.1pt,draw=black!15}
]

\addplot[
    draw=ourscolor,
    line width=1.2pt,
    mark=*,
    mark size=2.2pt,
    mark options={solid,fill=ourscolor,draw=ourscolor}
]
coordinates {
    (1,0.00)
    (2,0.03)
    (3,0.07)
    (4,0.17)
    (5,0.21)
};

\addplot[
    draw=oraclecolor,
    dashed,
    line width=1.0pt,
    mark=*,
    mark size=2.3pt,
    mark options={solid,fill=oraclecolor,draw=oraclecolor}
]
coordinates {
    (5,0.21)
    (6,0.20)
};

\end{groupplot}
\end{tikzpicture}

\caption{
\textbf{Controlled analysis on LIBERO-10 and expert-pool scaling.}
\textbf{(a)} Conditioning evidence.
\textbf{(b)} Bridge-reset ablation.
\textbf{(c)} Switching rule.
\textbf{(d)} Adapter source.
\textbf{(e)} Expert-pool scaling on CALVIN-D, reported as $\Delta$ Avg.\ relative to ABC-only training. Sources are added cumulatively; the final point additionally includes 149 DROID experts. Error bars denote 95\% bootstrap confidence intervals where available.
}
\vspace{-15pt}
\label{fig:libero_ablations}

\end{figure*}
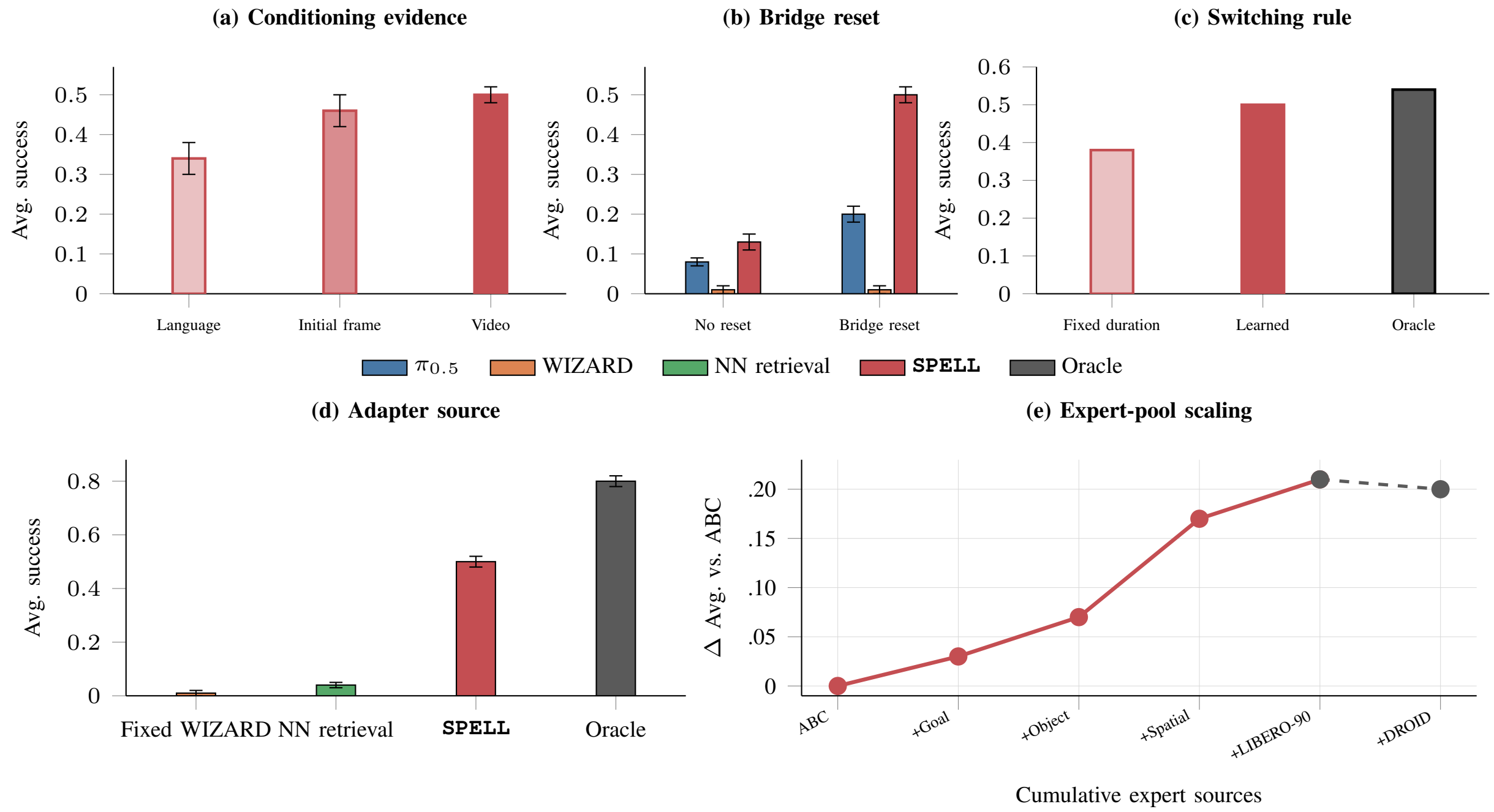

%% file: sections/6_lim_con.tex
\section{Conclusions}
We introduced \modelname, a sequential weight-space adaptation framework for long-horizon robotic manipulation. \modelname maps task evidence to skill-level adapters and dynamically composes them through termination-based switching. Experiments on CALVIN, LIBERO-10, Isaac Sim, and a real-world Franka Emika setup show improved long-horizon and state-dependent execution without target-task action supervision or test-time gradient-based adaptation. Across these settings, the gains become particularly evident as longer skill sequences are composed, supporting the use of dynamic weight-space specialization for sequential manipulation.

\noindent \textbf{Limitations.}
\modelname\ depends on the quality and diversity of the adapters used for meta-training, which limits performance on the most difficult compositions and under substantial state drift. The current method also assumes that the task is provided as an ordered sequence of subtasks; learning or predicting that decomposition is outside the present scope. Although termination-based switching approaches oracle switching in our LIBERO-10 ablation, execution can still fail when completion is predicted too early or too late. The bridge reset is specific to the LIBERO-10 protocol (CALVIN preserves scene and arm continuity) and may be undesirable in less structured deployments. Finally, the real-world experiments establish feasibility on one Franka setup rather than broad cross-embodiment generalization.